\documentclass[journal,twoside]{IEEEtran}
\usepackage{cite}
\usepackage{amsmath,amssymb,amsfonts}
\usepackage{algorithmic}
\usepackage{graphicx}
\usepackage{textcomp}
\usepackage{xcolor}
\usepackage{subcaption}
\usepackage{multirow}
\usepackage{hyperref}
\usepackage{diagbox}
\usepackage{graphicx,verbatim}
\usepackage{tikz}
\usepackage{booktabs} 
\usepackage{amssymb}
\usepackage{multirow}

\def\BibTeX{{\rm B\kern-.05em{\sc i\kern-.025em b}\kern-.08em
    T\kern-.1667em\lower.7ex\hbox{E}\kern-.125emX}}
\begin{document}
\title{D$^3$Seg: Dependency-Aware Diffusion for Brain Tumor Segmentation with Missing Modalities}
\author{Danish Ali, Ajmal Mian, Naveed Akhtar, and Ghulam Mubashar Hassan
\thanks{
This research is supported by the Australian Government
Research Training Scholarship. The authors also gratefully acknowledge the organizers of the BraTS2023 Challenge for providing the dataset used in this research. Dr. Naveed Akhtar is a recipient of the ARC Discovery Early Career Researcher Award (project \#DE230101058), funded by the Australian Government. }
\thanks{Danish Ali, Ajmal Mian, and Ghulam Mubashar Hassan are with The University of Western Australia, Perth, WA 6009, Australia (e-mail: danish.ali@research.uwa.edu.au; ajmal.mian@uwa.edu.au; ghulam.hassan@uwa.edu.au).}
\thanks{Naveed Akhtar is with The University of Melbourne, Melbourne, Parkville VIC 3010, Australia (e-mail: naveed.akhtar1@unimelb.edu.au).}
\thanks{Corresponding Author: Danish Ali.}
}

\maketitle

\begin{abstract}
Accurate brain tumor segmentation using multi-parametric MRI is critical for effective treatment planning. However, in clinical settings, complete acquisition of all MRI sequences is not always possible. The absence of certain MRI modalities results in substantial performance degradation in existing segmentation methods, which typically rely on naive feature concatenation or direct fusion strategies. To address this limitation, we propose a novel segmentation model D$^3$Seg which is designed to maintain stable performance under missing-modality settings. 
D$^3$Seg introduces Multi-hop Modality Graph Fusion (MMGF) to model higher-order inter-modality dependencies, a lightweight diffusion-based imputation mechanism to compensate for missing T1ce and FLAIR feature representations in latent space, and probability-space decision refinement to mitigate dominant-class overconfidence and improve delineation of underrepresented tumor subregions. We evaluate the proposed D$^3$Seg model on BraTS 2023 Glioma as the primary benchmark and further test it on a subset of the external BraTS 2023 Meningioma cohort to assess generalization across tumor pathologies.
The results are compared with the state-of-the-art models under different missing-modality conditions. The proposed model achieves approximately 1.5–2.0\% Dice improvement on enhancing tumor (ET) and around 1.0\% on tumor core (TC) across multiple missing-modality configurations compared to the current state-of-the-art model on BraTS Glioma dataset. Cross-cohort evaluation on BraTS Meningioma dataset demonstrates the generalizability of the proposed model, showing consistent improvements in the challenging TC and ET regions, with approximately 1.5-3.0\% and 1.5-6.5\% gains respectively across several missing-modality configurations.

\end{abstract}

\section{Introduction}
Brain tumors are among the most aggressive and life-threatening neurological diseases \cite{menze2014multimodal}. Precise delineation of tumor sub-regions is crucial for accurate diagnosis and effective treatment planning. Magnetic resonance imaging (MRI) is widely used to capture detailed structural information of brain tissue through multi-parametric image acquisition \cite{zhou2021latent}. Standard clinical protocols routinely acquire four MRI sequences \cite{bakas2017advancing}: T1, contrast-enhanced T1 (T1ce), T2 and FLAIR, which collectively provide complementary information required for delineating healthy and tumorous regions. However, the acquisition of all MRI modalities
is not always possible in clinical practice, as imaging protocols vary across institutions and contrast agents are not suitable for certain patients \cite{wang2021acn}.

Manual tumor segmentation is a labor-intensive process and is prone to inter-observer variability \cite{sharma2019automated}. Consequently, automatic brain tumor segmentation has emerged as an active area of research \cite{akil2020fully, wang2022relax}. However, most existing methods rely on complete multimodal MRI inputs \cite{havaei2017brain, she2023eoformer, ali2025attention, ali2025drbd, zhou2025bufnet, zhou2025dfuse}, and their performance degrades significantly in the absence of one or more MRI modalities \cite{zhu2025no}. To address this, studies have explored diverse modeling strategies to propose automatic brain tumor segmentation in the scenario of missing modalities \cite{liu2023sfusion,li2025dc,wang2023multi,shi2025fedamm,wang2025hypergraph}. These techniques differ in handling missing information and integrating it into the segmentation pipeline. 

One of the earliest approaches is HeMIS \cite{havaei2016hemis}, which introduces a hetero-modal segmentation framework that processes each MRI modality through a dedicated convolutional pipeline and aggregates the feature representations across available modalities using statistical operations. In contrast, Rob-Seg \cite{chen2019robust} proposes a feature disentanglement strategy combined with a gated feature fusion mechanism to improve robustness under missing-modality conditions. A2FSeg \cite{wang2023a2fseg} further extends this approach by introducing a two-stage fusion strategy that combines average feature aggregation with adaptive modality weighting to better exploit complementary information across modalities.
 
 Although fusion-based methods \cite{havaei2016hemis,wang2023a2fseg} have demonstrated promising performance, they primarily rely on information from the available modalities and lack explicit mechanisms to compensate for missing inputs \cite{zhou2022missing}. To address this limitation, reconstruction-based approaches have been proposed to synthesize missing modalities or learn multimodal representations by exploiting cross-modality correlations \cite{qi2025unified,zhang2025structure,zhou2020brain,zhou2021feature,xiong2025learning}. In this context, U-HVED \cite{dorent2019hetero} employs a hetero-modal variational encoder–decoder to learn a shared latent representation across modalities, jointly performing modality completion and segmentation. Similarly, M$^3$AE \cite{liu2023m3ae} proposes a masked autoencoding framework, where random subsets of MRI modalities and spatial regions of the remaining modalities are simultaneously masked, and the model learns to reconstruct the masked content. Building upon this masking strategy, M$^{3}$FeCon \cite{zeng2024missing} employs a multi-layer transformer to enable feature-to-feature reconstruction across arbitrary modality combinations. 
 However, reconstructing all missing modality features, irrespective of their contribution to the segmentation objective, may introduce unnecessary computational overhead and reduce practical efficiency \cite{zhu2025tackling}.

 Knowledge distillation-based methods \cite{wang2023learnable,zhu2025bridging} offer an alternative for segmentation under incomplete modality inputs by training student models corresponding to different missing-modality configurations under the supervision of a teacher model trained on complete modalities. However, the effectiveness of such designs is highly dependent on the 
 reliability of the teacher network, and they incur additional training overhead \cite{choi2023single}. 
 
 Beyond reconstruction and distillation-based methods, recent works have explored transformer and Mamba-based architectures \cite{dosovitskiyimage,gu2024mamba} to better capture long-range contextual dependencies \cite{xing2025segmamba,zhang2024tmformer}. Motivated by their success, IM-Fuse \cite{pipoli2025fuse} was recently introduced which adopts a Mamba-based backbone for multi-scale long-range contextual modeling, complemented by intra-modality and inter-modality transformer blocks at the bottleneck to refine cross-modality feature interactions. While this hierarchical global modeling improves whole tumor segmentation, performance gain for the clinically critical enhancing tumor (ET) remains inconsistent across different missing-modality configurations. Moreover, incorporating intra-modality transformer blocks increases computational cost.

 To overcome the above limitations, we propose D$^3$Seg, a dependency-aware diffusion-imputed segmentation network with decision refinement for 
 brain tumor segmentation with missing modalities. 
 Unlike existing fusion-based methods that rely solely on available modalities, or reconstruction-based approaches that indiscriminately synthesize all missing inputs, D$^3$Seg selectively imputes clinically critical modality information while explicitly modeling cross-modality dependencies and refining segmentation decisions in a targeted manner. 
 Our main contributions 
 are summarized below:
 \begin{enumerate}
    \item We propose a dependency-aware fusion approach that constructs
    a modality adjacency graph to model both direct and indirect inter-modality relationships via multi-hop propagation. This enables more expressive aggregation of inter-modality information compared to simple average fusion.
    \vspace{2mm}

    \item We propose a context-robust conditional latent diffusion module to directly synthesize the features of missing T1ce and FLAIR modalities at the network bottleneck. This provides targeted compensation for missing contrast information. 
    
    \vspace{2mm}
    \item We propose a unique decision refinement module that explicitly addresses dominant-class overconfidence under missing-modalities. 
    Our module adaptively redistributes probability mass toward underrepresented tumor subregions, improving subregion-level predictions and overall segmentation performance.
\end{enumerate}

Extensive evaluation on the BraTS 2023 Glioma \cite{baid2021rsna} and BraTS 2023 Meningioma \cite{labella2023asnr} datasets demonstrates that our method consistently improves segmentation performance compared to state-of-the-art approaches \cite{pipoli2025fuse,zeng2024missing,zhang2022mmformer,liu2023m3ae}. Specifically, the proposed method improves enhancing tumor (ET) Dice by approximately 1.5–2.0\% and tumor core (TC) Dice by around 1.0\% on the BraTS 2023 Glioma benchmark under multiple missing-modality configurations. On the external BraTS 2023 Meningioma cohort, the model further yields Dice gains of approximately 1.5–3.0\% in TC and 1.5–6.5\% in ET across several missing-modality settings, highlighting its effectiveness in clinically critical tumor subregions.

\section{Related Work}

This section reviews existing methods for brain tumor segmentation under missing MRI modality conditions, which are frequently encountered in clinical settings due to variations in acquisition protocols, scan quality, and patient conditions. Since segmentation models often rely on multiple complementary MRI sequences \cite{liu2023deep}, the absence of one or more modalities can substantially degrade performance. Existing approaches that address this problem can be broadly categorized into two main groups: modality synthesis methods, which attempt to reconstruct or generate the missing modalities from the available ones, and multi-task representation learning approaches, which aim to learn modality-robust representations that remain effective even when some modalities are unavailable.

\subsection{Modality Synthesis Methods}
Synthesis-based approaches aim to compensate for unavailable MRI sequences by inferring missing information from observed modalities before segmentation. These approaches mainly differ in methodology to recover missing information: in a shared latent feature space or directly in the image domain \cite{zhou2021latent,wu2023tiss,zhang2025structure, shaarawy2025modanet}. Latent-space synthesis methods, such as variational autoencoder-based approaches learn a shared representation across heterogeneous MRI modalities and use this representation to infer missing modality information \cite{zhu2021drm, dorent2025unified}. Image-domain synthesis methods instead generate the missing MRI sequence directly in the image space. Generative adversarial networks commonly follow image domain synthesis approach, where a generator predicts the absent modality and a discriminator encourages the synthesized image to resemble real MRI data \cite{sharma2019missing,conte2021generative,al2024flair}. Recently, diffusion-based methods have also been explored for missing MRI synthesis. Instead of relying on adversarial training, they estimate the target modality through an iterative denoising process. This denoising is performed either in the image domain \cite{meng2024multi, dorjsembe2024conditional} or in a latent feature space \cite{jiang2023cola, zhang2025structure}.

Although these approaches differ in their generative mechanisms, they share the same objective of exploiting cross-modality relationships among MRI sequences to approximate missing information and provide segmentation networks with enriched multimodal representations. By reconstructing absent sequences, they improve segmentation performance when clinically informative modalities are unavailable. However, explicit generation of full resolution MRI volumes introduces additional computational cost and may produce synthetic artifacts \cite{dayarathna2024deep,zhou2025stroke}. 

\subsection{Multi-task Representation Learning Methods}
Another approach to overcome missing modalities problem for brain tumor segmentation is to use
feature representations that remain stable when certain modalities are missing. Methods using this approach explore auxiliary supervision \cite{liang2025semantic}, knowledge transfer \cite{pani2024hybrid} and feature alignment \cite{zhuang20223d} which can be grouped as Multi-task representation learning methods. Recent approaches within this line of approach further incorporate transformer architectures to enhance long-range contextual modeling and cross-modal feature interaction under incomplete-modality conditions \cite{lu2023gmetanet, karimijafarbigloo2024mmcformer, konwer2023enhancing}. 

Multimodal medical transformer (mmFormer) \cite{zhang2022mmformer} introduces modality-specific hybrid encoders together with intra-modal and inter-modal transformer blocks to capture both modality-specific context and cross-modality relationships under incomplete-modality settings. It further employs auxiliary regularization in the encoder and decoder to improve robustness across arbitrary missing-modality combinations. Similarly, MMTSeg \cite{kang2026multimodal} combines Mamba and Transformer modules to enhance cross-modal feature learning and uses multimodal knowledge transfer to improve segmentation when only partial modality information is available. Despite their effectiveness, these methods often require careful balancing between the primary segmentation objective and auxiliary task learning, making performance sensitive to inter-task weight selection \cite{zhu2025xlstm}.

\section{Methodology}
\begin{figure*}[t]
\includegraphics[width=\textwidth]{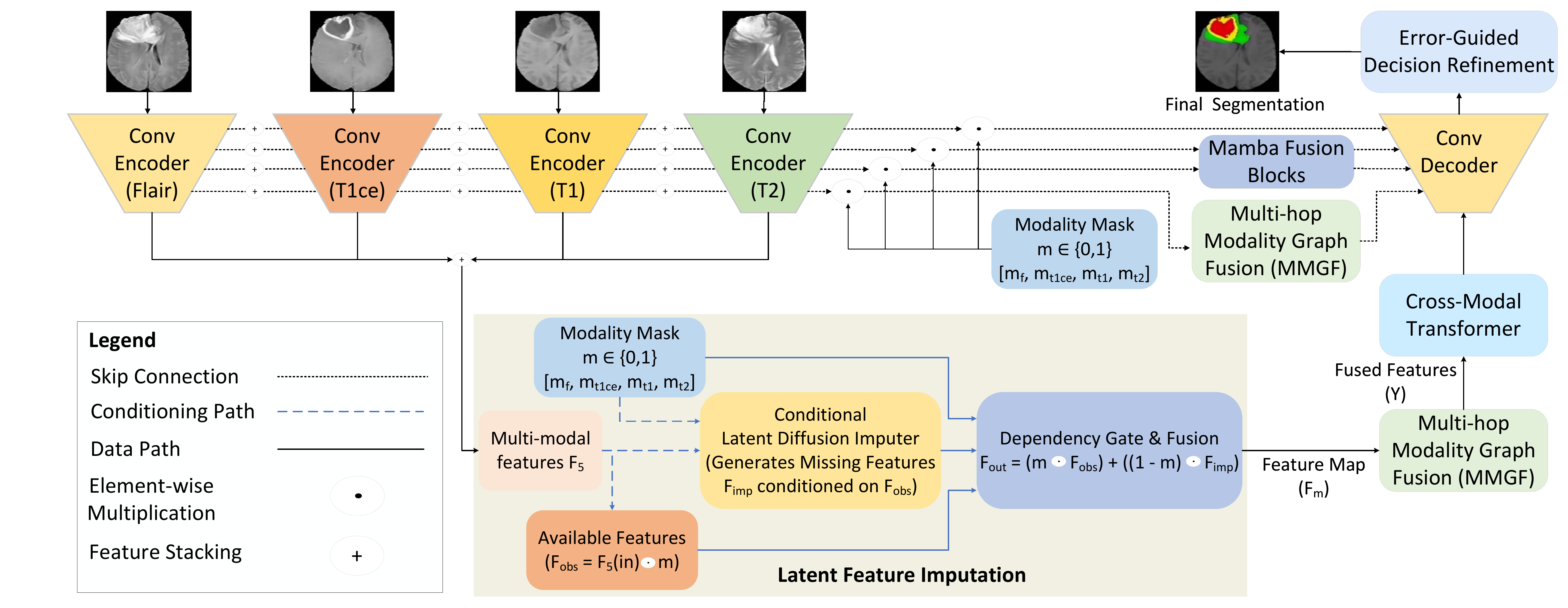}
\caption{Architecture of the proposed Dependency Aware Diffusion Imputed Decision Refined Segmentation (D$^3$Seg) network.} \label{archi}
\end{figure*}
The overall architecture of the proposed D$^3$Seg, a dependency-aware diffusion-imputed approach with decision refinement for brain tumor segmentation with missing modalities is presented in Fig.~\ref{archi}. 

Each MRI modality is processed by an independent 3D convolutional encoder to extract modality-specific feature representations. A binary modality mask ($m \in {0,1}$) is applied to MRI modality features, where $m=1$ for available modalities and $m=0$ for missing ones. The extracted features are adaptively integrated using a dependency-aware multi-hop modality graph fusion (MMGF) applied at the bottleneck and the skip connection of preceding encoder level. 

To compensate for missing contrast-enhanced and FLAIR information, a diffusion-based latent imputation module synthesizes T1ce and FLAIR modality features. The fused and imputed features are further processed by 
a cross-modal transformer at the bottleneck to capture global context. In addition, Mamba-based fusion blocks are incorporated into high-resolution skip connections to efficiently capture multi-scale long-range contextual dependencies while preserving fine-grained spatial details. The resulting features are decoded by a 3D convolutional decoder to produce an initial segmentation estimate. Finally, an error-guided decision refinement module adaptively redistributes class probability mass to improve the delineation of under-segmented tumor regions, particularly ET. The main components of D$^3$Seg are explained below.

\subsection{Multi-hop Modality Graph Fusion}
\label{mmgf}

\begin{figure*}[t]
\includegraphics[width=\textwidth]{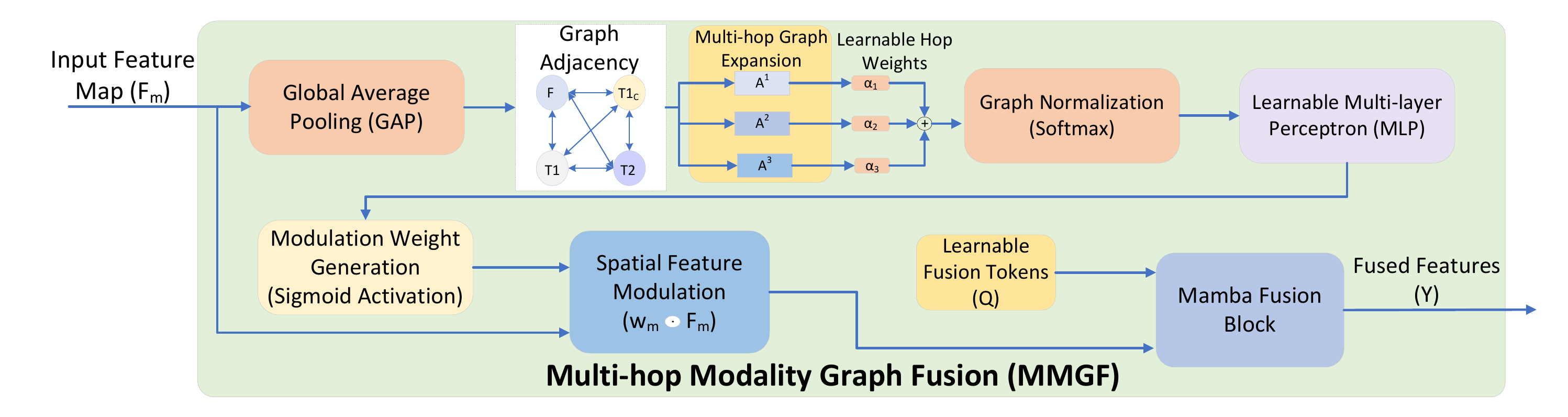}
\caption{Multi-hop modality graph fusion (MMGF) module for higher-order cross-modality feature interaction.
} \label{mmgffig}
\end{figure*}

Given heterogeneous and incomplete MRI modalities, effective fusion requires modeling inter-modality dependencies beyond simple aggregation. Therefore, we propose a Multi-hop Modality Graph Fusion (MMGF) module as presented in Fig.~\ref{mmgffig} to capture both direct and higher-order relationships among MRI sequences while preserving the spatial feature maps required for accurate tumor delineation.

Let $\mathbf{F}_m \in \mathbb{R}^{C \times D \times H \times W}$ denote the spatial feature map extracted from modality $m$, where $m \in \{1,\dots,M\}$ and $M$ is the total number of modalities. For each modality, we first apply global average pooling to obtain a compact modality descriptor:
\begin{equation}
\mathbf{h}_m = \operatorname{GAP}(\mathbf{F}_m),
\end{equation}
where $\mathbf{h}_m \in \mathbb{R}^{C}$ represents the global semantic descriptor of modality $m$. 
The pairwise dependency between modalities is encoded using a cosine-similarity-based adjacency matrix $A \in \mathbb{R}^{M \times M}$:

\begin{equation}
A =
\left[
\frac{
\mathbf{h}_i^\top \mathbf{h}_j
}{
\|\mathbf{h}_i\|_2 \|\mathbf{h}_j\|_2
}
\right]_{i,j=1}^{M},
\end{equation}
where $\mathbf{h}_i$ and $\mathbf{h}_j$ denote the modality descriptors of the $i$-th and $j$-th MRI sequences, respectively. When a modality is missing, the corresponding rows and columns of $A$ are masked to prevent unreliable interactions from unavailable inputs. For T1ce and FLAIR at the bottleneck level, the original modality features are used when available; otherwise, the proposed diffusion-based latent imputation module provides the corresponding imputed representations (details in \S~\ref{diff}).

To capture higher-order modality interactions, the masked adjacency matrix is expanded across multiple graph hops. This enables information from one modality to propagate through intermediate modalities rather than relying only on direct pairwise relationships. The normalized multi-hop adjacency matrix is computed as:

\begin{equation}
\hat{A}
=
\operatorname{softmax}
\left(
\sum_{k=1}^{3}
\alpha_k A^{(k)}
\right),
\end{equation}
where $A^{(k)}$ denotes the $k$-hop adjacency matrix and $\alpha_k$ are learnable hop weights that adaptively control the contribution of each graph-hop order. The graph-enhanced modality descriptors are then obtained by:

\begin{equation}
H^{\text{out}}
=
\phi
\left(
\hat{A}H
\right),
\end{equation}
where $H=[\mathbf{h}_1,\dots,\mathbf{h}_M]$ denotes the stacked modality descriptors and $\phi(\cdot)$ is a learnable multi-layer perceptron (MLP).

To preserve spatial information, MMGF does not directly use the pooled descriptors as segmentation features. Instead, the graph-enhanced modality descriptors are projected into modality-specific modulation weights, which are then applied to the original spatial feature maps:

\begin{equation}
\mathbf{w}_m
=
\sigma
\left(
\psi(\mathbf{h}^{\text{out}}_m)
\right), \qquad \mathbf{F}^{\text{out}}_m
=
\mathbf{w}_m \odot \mathbf{F}_m,
\end{equation}
where $\mathbf{h}^{\text{out}}_m$ is the graph-enhanced descriptor for modality $m$, $\psi(\cdot)$ is a learnable projection, $\sigma(\cdot)$ denotes the sigmoid activation function, and $\odot$ represents channel-wise modulation. This formulation preserves the original spatial feature map while adaptively enhancing or suppressing modality-specific responses according to learned inter-modality dependencies.

After graph-based modulation, the refined modality features are processed by a Mamba-based fusion block to obtain a compact high-level representation with spatial context. The graph-modulated features are arranged as a structured cross-modal token sequence, where learnable fusion tokens are introduced to capture complementary information from the modality-specific features. Through sequence modeling, the Mamba operator updates these fusion tokens by integrating cross-modal cues and propagating long-range contextual dependencies across the 3D spatial sequence.

\begin{equation}
\mathbf{Y}
=
\Gamma_{\mathrm{Mamba}}
\left(
\mathbf{F}^{\mathrm{out}}_1,
\dots,
\mathbf{F}^{\mathrm{out}}_M,
\mathbf{Q}
\right),
\end{equation}
where $\mathbf{Q}$ denotes the learnable fusion tokens, $\Gamma_{\mathrm{Mamba}}(\cdot)$ is the Mamba-based contextual fusion operator applied to the graph-modulated modality features and fusion tokens, and $\mathbf{Y}$ is the resulting sequence-level representation. The fusion tokens updated by Mamba sequence modeling are retained as the fused representation for subsequent high-level reasoning. In this design, the graph stage captures modality-level dependencies, whereas the Mamba stage consolidates the graph-refined features while preserving spatial structure and modeling long-range contextual interactions.

 The proposed MMGF is applied at the bottleneck and the immediately preceding encoder stage, where feature representations are more global and semantically rich compared to early encoder layers, making them better suited for modeling inter-modality dependencies.

\subsection{Context-Robust Conditional Latent Diffusion}
\label{diff}
\begin{figure*}[t]
\includegraphics[width=\textwidth]{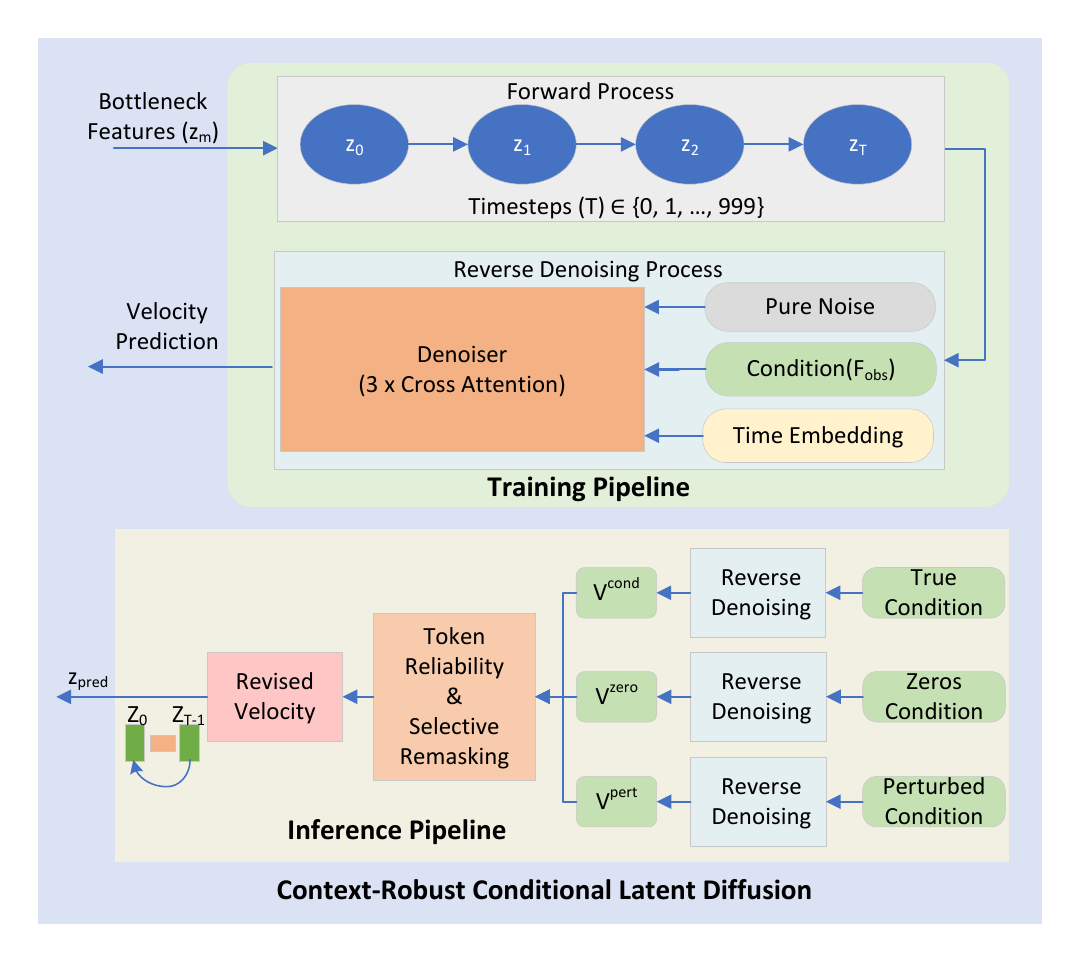}
\caption{Conditional latent diffusion imputer for T1ce and FLAIR latent feature imputation.} \label{diff_model}
\end{figure*}

Realizing the critical role of T1ce and FLAIR modalities in delineating tumor core and edema related boundaries, we propose a lightweight context robust conditional latent diffusion module to impute their features in the absence of these modalities as presented in Fig. \ref{diff_model}. 

Instead of relying on zero-filling, feature masking, or downstream fusion alone, the proposed module imputes missing T1ce and FLAIR features, using the available modalities as subject-specific guidance.
Diffusion is performed directly in the latent feature space of the T1ce and FLAIR encoder branches, where semantic information is compactly encoded and the sampling cost is substantially reduced. A lightweight transformer-based denoiser with three cross attention layers is used to remove injected Gaussian noise from the target latent representation, conditioned on the latent features extracted from the available modalities. The forward diffusion process for the bottleneck latent representation ($\mathbf{z}_{m}$) of the target missing modality $m \in \{\mathrm{T1ce}, \mathrm{FLAIR}\}$ is defined as:
\begin{equation}
\mathbf{z}_{t}
=
\alpha_t \mathbf{z}_{m}
+
\sigma_t \boldsymbol{\epsilon},
\qquad
\boldsymbol{\epsilon} \sim \mathcal{N}(\mathbf{0},\mathbf{I}),
\end{equation}
where $t$ is the diffusion timestep, and $\alpha_t$ and $\sigma_t$ define the noise schedule. We adopt the velocity parameterization to improve denoising stability across noise levels \cite{lin2024common}, with the target velocity defined as:
\begin{equation}
\mathbf{v}_{t}
=
\alpha_t \boldsymbol{\epsilon}
-
\sigma_t \mathbf{z}_{m}.
\end{equation}

The conditional denoiser $D_{\theta}$ takes the noisy target latent $\mathbf{z}_{t}$, timestep $t$, and conditioning representation $\mathbf{c}$ as input, and is optimized using the velocity-prediction objective:
\begin{equation}
\hat{\mathbf{v}}
=
D_{\theta}
\left(
\mathbf{z}_{t}, t, \mathbf{c}
\right),
\qquad
\mathcal{L}_{\mathrm{diff}}
=
\left\|
\mathbf{v}_{t}
-
\hat{\mathbf{v}}
\right\|_2^2,
\end{equation}
where $\hat{\mathbf{v}}$ denotes the predicted velocity. By conditioning the denoising process on $\mathbf{c}$, the model learns to impute the target latent representation using subject-specific information from the available MRI modalities. The same conditional diffusion formulation is used for both T1ce and FLAIR imputation by changing the target latent. 

During inference, sampling starts from random noise and progressively reconstructs the missing latent feature under the available-modality condition. However, conditional diffusion can still produce locally unreliable latent tokens when the denoising prediction is overly sensitive to small perturbations in the conditioning representation \cite{zhai2026core}. To improve conditional reliability, we incorporate a context-robust remasking strategy during sampling.

At selected denoising steps, the denoiser is evaluated under three conditioning as shown in Fig. \ref{diff_model}: the true condition $\mathbf{c}$, a zero condition $\mathbf{0}$, and a perturbed condition $\tilde{\mathbf{c}}$. The perturbed condition is obtained by randomly masking a subset of conditioning tokens:
\begin{equation}
\tilde{\mathbf{c}}
=
\mathbf{m}_{c}
\odot
\mathbf{c},
\end{equation}
where $\mathbf{m}_{c} \in \{0,1\}^{N \times 1}$ is a random token mask and $\odot$ denotes token-wise multiplication. This produces three velocity predictions:
\begin{equation}
\hat{\mathbf{v}}^{\mathrm{cond}}
=
D_{\theta}(\mathbf{z}_{t},t,\mathbf{c}),
\quad
\hat{\mathbf{v}}^{\mathrm{zero}}
=
D_{\theta}(\mathbf{z}_{t},t,\mathbf{0}),
\quad
\hat{\mathbf{v}}^{\mathrm{pert}}
=
D_{\theta}(\mathbf{z}_{t},t,\tilde{\mathbf{c}}).
\end{equation}
Token-wise reliability is estimated by comparing context instability with condition contribution:
\begin{equation}
r_i
=
\frac{
\left\|
\hat{\mathbf{v}}^{\mathrm{cond}}_i
-
\hat{\mathbf{v}}^{\mathrm{pert}}_i
\right\|_2^2
}{
\left\|
\hat{\mathbf{v}}^{\mathrm{cond}}_i
-
\hat{\mathbf{v}}^{\mathrm{zero}}_i
\right\|_2^2
+
\varepsilon
},
\end{equation}
where the numerator measures how sensitive token $i$ is to perturbations in the conditioning context, while the denominator measures the strength of the conditional signal relative to an unconditioned prediction. 
A high value of $r_i$ indicates that the prediction at token $i$ is disproportionately sensitive to perturbations of the conditioning context relative to its estimated conditional contribution. We therefore treat high-scoring tokens as unreliable conditional predictions and select them for remasking and correction, while preserving lower-scoring tokens to avoid unnecessary changes to stable latent regions.

For the selected unreliable token set $\Omega$, we partially re-noise the current latent state using the initial sampling noise:
\begin{equation}
\tilde{\mathbf{z}}_{t,i}
=
\begin{cases}
\lambda \mathbf{z}_{t,i} + (1-\lambda)\boldsymbol{\eta}_{i}, & i \in \Omega, \\
\mathbf{z}_{t,i}, & i \notin \Omega,
\end{cases}
\end{equation}
where $\boldsymbol{\eta}$ denotes the initial Gaussian noise and $\lambda$ controls the remasking strength. The denoiser is then re-evaluated on the remasked latent using the true condition:
\begin{equation}
\hat{\mathbf{v}}^{\mathrm{rev}}
=
D_{\theta}
\left(
\tilde{\mathbf{z}}_{t},t,\mathbf{c}
\right),
\end{equation}
where $\hat{\mathbf{v}}^{\mathrm{rev}}$ denotes the revised velocity prediction obtained after remasking and denoising the remasked latent. The final velocity prediction for the current sampling step is obtained by replacing only the selected unreliable token predictions:
\begin{equation}
\hat{\mathbf{v}}^{\mathrm{final}}_i
=
\begin{cases}
\hat{\mathbf{v}}^{\mathrm{rev}}_i, & i \in \Omega, \\
\hat{\mathbf{v}}^{\mathrm{cond}}_i, & i \notin \Omega.
\end{cases}
\end{equation}
The scheduler then updates the latent sample using $\hat{\mathbf{v}}^{\mathrm{final}}$. 

In our implementation, context-robust remasking is applied only during the intermediate sampling steps. At this stage, unreliable token predictions can be identified from their conditional sensitivity. This avoids unnecessary perturbation during the early generation phase and preserves the stability of the final refinement steps. The final sampled latent representation is inserted into the segmentation pipeline as the imputed T1ce and FLAIR features before MMGF. This provides the subsequent graph-based modulation and Mamba-based fusion stages with T1ce and FLAIR related bottleneck representations when these critical modalities are unavailable, enabling modality-aware reasoning under incomplete MRI inputs.
\begin{table*}[t]
\setlength{\tabcolsep}{3pt}
\newcommand{\pcirc}[1]{%
\tikz[baseline=(n.base)]{
\node[
  rectangle,
  draw,
  fill=gray!35,
  minimum size=7mm,
  inner sep=0pt,
  align=center
] (n) {\strut #1};
}}

\newcommand{\mcirc}[1]{%
\tikz[baseline=(n.base)]{
\node[
  rectangle,
  draw,
  fill=white,
  minimum size=7mm,
  inner sep=0pt,
  align=center
] (n) {\strut #1};
}}

\centering
\caption{Quantitative results on BraTS 2023 Glioma across different missing modalities. Each column corresponds to a specific missing-modality configuration. Gray squares indicate available MRI modalities, whereas white squares denote missing modalities. Dice Score (\%) is reported for Whole Tumor (WT), Tumor Core (TC), and Enhancing Tumor (ET). Bold and underlined values represent the best and second-best results.}
\label{tab:brats2023_all}

{
\resizebox{\textwidth}{!}{%
\begin{tabular}{l|ccccccccccccccc}
\hline
\multirow{4}{*}{Models}& \multicolumn{15}{c}{Modality Configurations} \\
\cline{2-16}
 &

\begin{tabular}{@{}c@{}c@{}}\pcirc{F}&\mcirc{T1}\\\mcirc{T1c}&\mcirc{T2}\end{tabular} &
\begin{tabular}{@{}c@{}c@{}}\mcirc{F}&\pcirc{T1}\\\mcirc{T1c}&\mcirc{T2}\end{tabular} &
\begin{tabular}{@{}c@{}c@{}}\mcirc{F}&\mcirc{T1}\\\pcirc{T1c}&\mcirc{T2}\end{tabular} &
\begin{tabular}{@{}c@{}c@{}}\mcirc{F}&\mcirc{T1}\\\mcirc{T1c}&\pcirc{T2}\end{tabular} &
\begin{tabular}{@{}c@{}c@{}}\pcirc{F}&\pcirc{T1}\\\mcirc{T1c}&\mcirc{T2}\end{tabular} &
\begin{tabular}{@{}c@{}c@{}}\pcirc{F}&\mcirc{T1}\\\pcirc{T1c}&\mcirc{T2}\end{tabular} &
\begin{tabular}{@{}c@{}c@{}}\pcirc{F}&\mcirc{T1}\\\mcirc{T1c}&\pcirc{T2}\end{tabular} &
\begin{tabular}{@{}c@{}c@{}}\mcirc{F}&\pcirc{T1}\\\pcirc{T1c}&\mcirc{T2}\end{tabular} &
\begin{tabular}{@{}c@{}c@{}}\mcirc{F}&\pcirc{T1}\\\mcirc{T1c}&\pcirc{T2}\end{tabular} &
\begin{tabular}{@{}c@{}c@{}}\mcirc{F}&\mcirc{T1}\\\pcirc{T1c}&\pcirc{T2}\end{tabular} &
\begin{tabular}{@{}c@{}c@{}}\pcirc{F}&\pcirc{T1}\\\pcirc{T1c}&\mcirc{T2}\end{tabular} &
\begin{tabular}{@{}c@{}c@{}}\pcirc{F}&\pcirc{T1}\\\mcirc{T1c}&\pcirc{T2}\end{tabular} &
\begin{tabular}{@{}c@{}c@{}}\pcirc{F}&\mcirc{T1}\\\pcirc{T1c}&\pcirc{T2}\end{tabular} &
\begin{tabular}{@{}c@{}c@{}}\mcirc{F}&\pcirc{T1}\\\pcirc{T1c}&\pcirc{T2}\end{tabular} &
\begin{tabular}{@{}c@{}c@{}}\pcirc{F}&\pcirc{T1}\\\pcirc{T1c}&\pcirc{T2}\end{tabular} \\
\hline

\multicolumn{16}{c}{{Whole Tumor Dice Score (\%)}} \\
\hline


mmForm \cite{zhang2022mmformer}  & 91.4 & \underline{82.8} & \textbf{83.7} & \underline{88.5} & 92.2 & \underline{92.7} & 91.3 & \textbf{85.5} & \underline{89.8} & \underline{90.1} & \textbf{92.8} & 92.5 & \underline{93.0} & \textbf{90.5} & 93.0 \\

SFusion \cite{liu2023sfusion}  & 89.1 & 78.5 & 77.6 & 87.0 & 90.8 & 91.2 & 91.3 & 81.3 & 88.2 & 88.7 & 91.7 & 91.6 & 92.1 & 88.8 & 92.2 \\

ShaSpec \cite{wang2023multi}  & 91.0 & 79.9 & 79.5 & 86.9 & 91.9 & 92.3 & 92.2 & 82.7 & 88.3 & 88.7 & \underline{92.6} & 92.5 & 92.9 & 89.2 & 93.0 \\

$M^3$AE \cite{liu2023m3ae}     & \underline{91.5} & 81.7 & 82.5 & \underline{88.5} & 91.9 & 92.5 & 92.2 & 83.6 & 89.2 & 89.7 & \underline{92.6} & 92.1 & \underline{93.0} & 90.0 & 92.9 \\

$M^3$FeCon \cite{zeng2024missing}  & 87.7 & 81.2 & 81.2 & \underline{88.5} & 89.4 & 90.0 & 92.1 & 83.9 & \underline{89.8} & 89.5 & 90.5 & 92.7 & 92.6 & 90.1 & \underline{93.1} \\

IM-Fuse \cite{pipoli2025fuse}
         & \textbf{91.8} & \textbf{83.0} & \textbf{83.7} & \textbf{88.7} & \underline{92.4} & \textbf{92.8} & \underline{92.6} & \textbf{85.5} & \textbf{90.1} & \textbf{90.2} & \textbf{92.8} & \textbf{93.0} & \textbf{93.1} & \textbf{90.5} & \textbf{93.3} \\

Proposed & \textbf{91.8} & 82.5 & \underline{83.6} & \textbf{88.7} & \textbf{92.6} & \textbf{92.8} & \textbf{92.8} & \underline{85.3} & 89.8 & \textbf{90.2} & \textbf{92.8} & \underline{92.8} & \underline{93.0} & \underline{90.4} & 93.0 \\
\hline

\multicolumn{16}{c}{{Tumor Core Dice Score (\%)}} \\
\hline


mmForm \cite{zhang2022mmformer}  & 78.3 & 73.6 & 89.2 & 74.5 & 80.6 & 90.7 & \underline{80.0} & 90.1 & 77.5 & 90.6 & 90.9 & 80.8 & 91.0 & 90.8 & 91.0 \\

SFusion \cite{liu2023sfusion}  & 74.0 & 70.4 & 86.7 & 74.3 & 77.4 & 88.6 & 77.8 & 88.4 & 76.5 & 89.4 & 89.1 & 78.5 & 89.3 & 89.5 & 89.5 \\

ShaSpec \cite{wang2023multi}  & 74.3 & 71.5 & 87.8 & 72.6 & 78.1 & 89.7 & 77.3 & 89.3 & 76.2 & 89.6 & 90.3 & 79.1 & 90.4 & 90.0 & 90.7 \\

$M^3$AE \cite{liu2023m3ae}     & 76.8 & \underline{75.9} & 89.9 & \textbf{77.9} & 79.9 & \underline{90.9} & 79.2 & 90.5 & 78.9 & \underline{90.8} & \underline{91.2} & 79.9 & \textbf{91.5} & 91.0 & \textbf{91.5} \\

$M^3$FeCon \cite{zeng2024missing}  & 72.3 & 74.1 & \underline{90.1} & 75.9 & 77.1 & \underline{90.9} & 79.1 & \underline{90.6} & \underline{79.5} & 90.7 & \underline{91.2} & 80.9 & 91.2 & \underline{91.4} & \underline{91.1} \\

IM-Fuse \cite{pipoli2025fuse}
         & \underline{78.8} & 75.4 & \textbf{90.5} & 76.5 & \textbf{80.9} & \textbf{91.4} & 79.9 & \textbf{91.2} & 79.1 & \textbf{91.2} & \textbf{91.6} & \underline{81.4} & \underline{91.3} & \textbf{91.5} & \textbf{91.5} \\

Proposed & \textbf{79.1} & \textbf{76.6} & 89.5 & \underline{77.7} & \underline{80.8} & 90.8 & \textbf{81.4} & 90.3 & \textbf{80.5} & 90.7 & 91.0 & \textbf{81.9} & 91.1 & 90.8 & \underline{91.1} \\
\hline
\multicolumn{16}{c}{{Enhancing Tumor Dice Score (\%)}} \\
\hline


mmForm \cite{zhang2022mmformer}  & 58.8 & 54.7 & \underline{84.1} & 58.3 & 62.5 & 84.9 & 63.9 & 84.7 & 62.5 & 84.7 & 84.8 & 66.0 & 84.2 & \underline{85.9} & 84.7 \\

SFusion \cite{liu2023sfusion}  & 52.2 & 48.9 & 82.2 & 54.3 & 57.4 & 83.9 & 59.0 & 83.6 & 57.1 & 83.8 & 83.8 & 60.1 & 83.9 & 84.0 & 84.0 \\

ShaSpec \cite{wang2023multi}  & 53.3 & 49.1 & 80.5 & 52.4 & 57.4 & 81.9 & 58.0 & 81.7 & 56.1 & 81.9 & 82.4 & 59.6 & 82.1 & 82.4 & 82.4 \\

$M^3$AE \cite{liu2023m3ae}     & 56.7 & 56.0 & 82.5 & 58.8 & 60.7 & \underline{85.6} & 61.2 & \underline{84.9} & 60.6 & \underline{85.8} & \underline{85.9} & 62.3 & \underline{85.7} & 85.1 & 85.6 \\

$M^3$FeCon \cite{zeng2024missing}  & 53.2 & 53.8 & 82.8 & 56.6 & 58.0 & 83.5 & 60.4 & 84.3 & 61.2 & 83.9 & 84.0 & 62.4 & 84.0 & 84.4 & 84.2 \\

IM-Fuse \cite{pipoli2025fuse}
         & \underline{59.5} & \underline{56.3} & 83.5 & \underline{59.6} & \underline{63.9} & 85.0 & \underline{64.3} & 84.5 & \underline{63.6} & 84.9 & 85.1 & \underline{67.0} & 85.2 & 85.6 & \underline{85.8} \\

Proposed & \textbf{60.6} & \textbf{59.3} & \textbf{86.4} & \textbf{61.2} & \textbf{65.9} & \textbf{87.3} & \textbf{67.3} & \textbf{86.9} & \textbf{65.6} & \textbf{87.3} & \textbf{87.5} & \textbf{68.7} & \textbf{87.5} & \textbf{87.3} & \textbf{87.4} \\
\hline
\end{tabular}}
}
\end{table*}
\subsection{Error-Guided Decision Refinement}
The absence of certain MRI modalities often leads to under-segmentation of minority tumor classes, particularly the enhancing tumor, together with an overconfidence bias toward dominant classes such as edema (ED). To address this, we propose a unique error-guided decision refinement (EGDR) module that operates directly in the probability space to refine error-prone predictions. 

Given the decoder logits, a lightweight multi-scale convolution-based error predictor estimates a voxel-wise error likelihood map by aggregating local and dilated contextual information. This highlights the regions prone to under-segmentation for enhancing tumor. The predicted error likelihood map is used to adaptively redistribute probability mass between related tumor classes:
\begin{equation}
P'_{et} = P_{et} + w_{et} \odot \Delta; \quad
P'_{ed} = P_{ed} - w_{ed} \odot \Delta; \qquad
\Delta = e \odot P_{ed},
\end{equation}
where \(e\) denotes the voxel-wise error likelihood map, \(\odot\) represents element-wise multiplication, and \(w_{et}\) and \(w_{ed}\) are learnable weights that control the amount of probability mass $(\Delta)$ transferred from ED to ET. The refined probabilities \((P'_{et}, P'_{ed})\) are normalized to preserve a valid distribution. The EGDR design increases ET confidence in under-segmented regions while suppressing ED overconfidence, resulting in balanced tumor subregion predictions.

\section{Experiments}
We evaluate the proposed method on the BraTS 2023 Glioma benchmark \cite{baid2021rsna}, which contains 1,251 multi-parametric brain MRI scans with expert-annotated tumor labels. Each subject includes the standard MRI modalities used in brain tumor segmentation, and the corresponding annotations. Following recent missing-modality segmentation literature \cite{pipoli2025fuse}, we split the dataset into 70\%, 10\%, and 20\% for training, validation, and testing, respectively. All splits are performed at the subject level to avoid data leakage.

In addition to the BraTS 2023 Glioma benchmark, we further assess the generalization ability of the proposed model and competing state of the art methods on an external subset of the BraTS 2023 Meningioma dataset. This external evaluation is particularly important because most existing missing-modality brain tumor segmentation studies are evaluated only on BraTS Glioma or on variants derived from previous BraTS Glioma challenges. As a result, their reported performance may not fully reflect robustness to tumor-type shift. To the best of our knowledge, validation beyond the glioma domain remains largely unexplored in the missing-modality brain tumor segmentation setting.

To address this gap, we construct an external test cohort of 279 cases from the BraTS Meningioma dataset. To ensure a clinically meaningful and fair evaluation across all missing-modality scenarios, we include only cases in which all three tumor sub-regions are present: edema, enhancing tumor, and non-enhancing tumor. This selection criterion is important because different MRI modalities provide complementary information for different tumor components; therefore, evaluating cases containing all tumor sub-regions allows meaningful assessment of model robustness when one or more modalities are missing. The BraTS Meningioma cohort is used strictly for external testing and is not involved in training, validation, hyperparameter tuning, or model selection.
\begin{table*}[t]
\setlength{\tabcolsep}{1pt}
\newcommand{\pcirc}[1]{%
\tikz[baseline=(n.base)]{
\node[
  rectangle,
  draw,
  fill=gray!35,
  minimum size=7mm,
  inner sep=0pt,
  align=center
] (n) {\strut #1};
}}

\newcommand{\mcirc}[1]{%
\tikz[baseline=(n.base)]{
\node[
  rectangle,
  draw,
  fill=white,
  minimum size=7mm,
  inner sep=0pt,
  align=center
] (n) {\strut #1};
}}

\centering
\caption{Quantitative results on BraTS 2023 Meningioma across different missing modalities. Each column corresponds to a specific missing-modality configuration. Gray squares indicate available MRI modalities, whereas white squares denote missing modalities. Dice Score (\%) is reported for Whole Tumor (WT), Tumor Core (TC), and Enhancing Tumor (ET). Bold and underlined values represent the best and second-best results.}
\label{tab:brats2023_men}

{
\setlength{\tabcolsep}{3pt}
\resizebox{\textwidth}{!}{%
\begin{tabular}{l|ccccccccccccccc}
\hline
\multirow{4}{*}{Models}& \multicolumn{15}{c}{Modality Configurations}\\
\cline{2-16}
 &

\begin{tabular}{@{}c@{}c@{}}\pcirc{F}&\mcirc{T1}\\\mcirc{T1c}&\mcirc{T2}\end{tabular} &
\begin{tabular}{@{}c@{}c@{}}\mcirc{F}&\pcirc{T1}\\\mcirc{T1c}&\mcirc{T2}\end{tabular} &
\begin{tabular}{@{}c@{}c@{}}\mcirc{F}&\mcirc{T1}\\\pcirc{T1c}&\mcirc{T2}\end{tabular} &
\begin{tabular}{@{}c@{}c@{}}\mcirc{F}&\mcirc{T1}\\\mcirc{T1c}&\pcirc{T2}\end{tabular} &
\begin{tabular}{@{}c@{}c@{}}\pcirc{F}&\pcirc{T1}\\\mcirc{T1c}&\mcirc{T2}\end{tabular} &
\begin{tabular}{@{}c@{}c@{}}\pcirc{F}&\mcirc{T1}\\\pcirc{T1c}&\mcirc{T2}\end{tabular} &
\begin{tabular}{@{}c@{}c@{}}\pcirc{F}&\mcirc{T1}\\\mcirc{T1c}&\pcirc{T2}\end{tabular} &
\begin{tabular}{@{}c@{}c@{}}\mcirc{F}&\pcirc{T1}\\\pcirc{T1c}&\mcirc{T2}\end{tabular} &
\begin{tabular}{@{}c@{}c@{}}\mcirc{F}&\pcirc{T1}\\\mcirc{T1c}&\pcirc{T2}\end{tabular} &
\begin{tabular}{@{}c@{}c@{}}\mcirc{F}&\mcirc{T1}\\\pcirc{T1c}&\pcirc{T2}\end{tabular} &
\begin{tabular}{@{}c@{}c@{}}\pcirc{F}&\pcirc{T1}\\\pcirc{T1c}&\mcirc{T2}\end{tabular} &
\begin{tabular}{@{}c@{}c@{}}\pcirc{F}&\pcirc{T1}\\\mcirc{T1c}&\pcirc{T2}\end{tabular} &
\begin{tabular}{@{}c@{}c@{}}\pcirc{F}&\mcirc{T1}\\\pcirc{T1c}&\pcirc{T2}\end{tabular} &
\begin{tabular}{@{}c@{}c@{}}\mcirc{F}&\pcirc{T1}\\\pcirc{T1c}&\pcirc{T2}\end{tabular} &
\begin{tabular}{@{}c@{}c@{}}\pcirc{F}&\pcirc{T1}\\\pcirc{T1c}&\pcirc{T2}\end{tabular} \\
\hline


\multicolumn{16}{c}{{Whole Tumor Dice Score (\%)}} \\
\hline
mmForm \cite{zhang2022mmformer}
& \textbf{82.7} & 74.3 & \underline{81.2} & \textbf{78.8} & \underline{84.9} & 88.1 & \textbf{83.6} & 82.5 & \underline{82.0} & 86.6 & 88.4 & \textbf{84.9} & 88.3 & 86.8 & 88.5 \\

IM-Fuse \cite{pipoli2025fuse}
& \underline{81.8} & \textbf{75.7} & \underline{81.2} & 78.1 & \textbf{85.1} & \underline{88.7} & \textbf{83.6} & \underline{82.7} & \textbf{82.1} & \underline{86.9} & \underline{89.1} & \textbf{84.9} & \underline{89.0} & \underline{87.1} & \underline{89.2} \\

Proposed
& 81.7 & \underline{75.0} & \textbf{81.3} & \underline{78.7} & 84.4 & \textbf{89.6} & \underline{83.1} & \textbf{82.9} & 81.7 & \textbf{87.9} & \textbf{89.7} & \underline{84.1} & \textbf{89.7} & \textbf{87.7} & \textbf{89.7} \\
\hline

\multicolumn{16}{c}{{Tumor Core Dice Score (\%)}} \\
\hline
mmForm \cite{zhang2022mmformer}
& 62.9 & 58.1 & 90.6 & 54.6 & \underline{66.2} & 90.1 & 63.5 & \underline{92.1} & 58.7 & \underline{91.2} & 91.1 & 65.3 & 90.1 & \underline{91.8} & \underline{90.7} \\

IM-Fuse \cite{pipoli2025fuse}
& \underline{63.6} & \underline{61.6} & \underline{90.8} & \underline{58.7} & \textbf{69.2} & \underline{90.9} & \underline{65.4} & 91.4 & \underline{65.2} & 90.8 & \underline{91.4} & \underline{67.8} & \underline{90.4} & 91.1 & \underline{90.7} \\

Proposed
& \textbf{66.1} & \textbf{64.9} & \textbf{92.2} & \textbf{64.1} & \textbf{69.2} & \textbf{92.8} & \textbf{68.0} & \textbf{92.6} & \textbf{67.0} & \textbf{92.9} & \textbf{93.1} & \textbf{69.3} & \textbf{92.8} & \textbf{92.9} & \textbf{92.9} \\
\hline

\multicolumn{16}{c}{{Enhancing Tumor Dice Score (\%)}} \\
\hline
mmForm \cite{zhang2022mmformer}
& 43.8 & 41.9 & \underline{90.0} & 46.3 & 49.2 & 89.6 & 52.4 & \underline{91.3} & 49.6 & \underline{90.4} & \underline{90.6} & 54.1 & \underline{89.5} & \underline{91.0} & \underline{90.1} \\

IM-Fuse \cite{pipoli2025fuse}
& \textbf{49.2} & \textbf{50.3} & 89.9 & \underline{48.3} & \textbf{56.7} & \underline{89.8} & \underline{54.4} & 90.3 & \underline{55.8} & 89.5 & 90.3 & \underline{57.2} & 89.2 & 89.9 & 89.7 \\

Proposed & \underline{45.8} & \underline{48.1} & \textbf{91.2} & \textbf{55.6} & \underline{54.0} & \textbf{92.0} & \textbf{57.2} & \textbf{91.5} & \textbf{58.5} & \textbf{91.9} & \textbf{92.1} & \textbf{57.3} & \textbf{91.9} & \textbf{92.0} & \textbf{92.1} \\
\hline
\end{tabular}}
}
\end{table*}
Segmentation performance of the proposed model for both datasets is evaluated using the Dice similarity coefficient for the standard clinically relevant tumor regions: whole tumor (WT), tumor core (TC), and enhancing tumor (ET). These regions capture complementary anatomical aspects of the tumor and are widely used in BraTS-based segmentation evaluation.

The proposed model is implemented in PyTorch and trained for 1,000 epochs using the Adam optimizer with an initial learning rate of (2$\times10^{-4}$). All experiments are conducted on an NVIDIA RTX 3090 GPU with a batch size of 3. To improve generalization and reduce the risk of overfitting, we apply data augmentation during training, including random flipping, random rotations, and intensity-based perturbations such as intensity shifting and intensity scaling. 
\begin{table*}[!h]
\setlength{\tabcolsep}{14pt}
\newcommand{\pcirc}[1]{%
\tikz[baseline=(n.base)]{
\node[
  rectangle,
  draw,
  fill=gray!35,
  minimum size=7mm,
  inner sep=0pt,
  align=center
] (n) {\strut #1};
}}

\newcommand{\mcirc}[1]{%
\tikz[baseline=(n.base)]{
\node[
  rectangle,
  draw,
  fill=white,
  minimum size=7mm,
  inner sep=0pt,
  align=center
] (n) {\strut #1};
}}

\newcommand{\modcfg}[4]{%
\begin{tabular}{@{}c@{}c@{\hspace{0mm}}c@{}c@{}}
#1 & #2 & #3 & #4
\vspace{1mm}
\end{tabular}
}

\centering
\caption{Detailed statistical comparison between the proposed method and IM-Fuse on BraTS 2023 Meningioma. Paired two-sided Wilcoxon signed-rank tests were performed using case-wise Dice scores under identical modality configurations. Reported values are raw $p$ and FDR-corrected $p_{\mathrm{FDR}}$ values.}
\label{tab:brats2023_full_stat_results}

{
\renewcommand{\arraystretch}{1.5}
\resizebox{\textwidth}{!}{%
\begin{tabular}{c|cc|cc|cc}
\hline
\multirow{2}{*}{Configuration} 
& \multicolumn{2}{c|}{WT} 
& \multicolumn{2}{c|}{TC} 
& \multicolumn{2}{c}{ET} \\
\cline{2-7}
& $p$ & $p_{\mathrm{FDR}}$ 
& $p$ & $p_{\mathrm{FDR}}$ 
& $p$ & $p_{\mathrm{FDR}}$ \\
\hline
\modcfg{\pcirc{F}}{\mcirc{T1}}{\mcirc{T1c}}{\mcirc{T2}}
& 0.2336 & 0.2695 & 0.0008 & 0.0011 & 0.0003 & 0.0004 \\

\modcfg{\mcirc{F}}{\mcirc{T1}}{\pcirc{T1c}}{\mcirc{T2}}
& 0.4920 & 0.5031 & $<0.0001$ & $<0.0001$ & $<0.0001$ & $<0.0001$ \\

\modcfg{\mcirc{F}}{\pcirc{T1}}{\mcirc{T1c}}{\mcirc{T2}}
& 0.0018 & 0.0024 & 0.0006 & 0.0009 & 0.0583 & 0.0691 \\

\modcfg{\mcirc{F}}{\mcirc{T1}}{\mcirc{T1c}}{\pcirc{T2}}
& 0.0124 & 0.0155 & $<0.0001$ & $<0.0001$ & $<0.0001$ & $<0.0001$ \\

\modcfg{\mcirc{F}}{\mcirc{T1}}{\pcirc{T1c}}{\pcirc{T2}}
& $<0.0001$ & $<0.0001$ & $<0.0001$ & $<0.0001$ & $<0.0001$ & $<0.0001$ \\

\modcfg{\mcirc{F}}{\pcirc{T1}}{\pcirc{T1c}}{\mcirc{T2}}
& 0.3911 & 0.4190 & $<0.0001$ & $<0.0001$ & $<0.0001$ & $<0.0001$ \\

\modcfg{\pcirc{F}}{\pcirc{T1}}{\mcirc{T1c}}{\mcirc{T2}}
& 0.2616 & 0.2943 & 0.7732 & 0.7732 & 0.0001 & 0.0002 \\

\modcfg{\mcirc{F}}{\pcirc{T1}}{\mcirc{T1c}}{\pcirc{T2}}
& 0.2897 & 0.3179 & 0.0121 & 0.0155 & 0.0001 & 0.0002 \\

\modcfg{\pcirc{F}}{\mcirc{T1}}{\mcirc{T1c}}{\pcirc{T2}}
& $<0.0001$ & $<0.0001$ & 0.0001 & 0.0002 & 0.0002 & 0.0003 \\

\modcfg{\pcirc{F}}{\mcirc{T1}}{\pcirc{T1c}}{\mcirc{T2}}
& $<0.0001$ & $<0.0001$ & $<0.0001$ & $<0.0001$ & $<0.0001$ & $<0.0001$ \\

\modcfg{\pcirc{F}}{\pcirc{T1}}{\pcirc{T1c}}{\mcirc{T2}}
& $<0.0001$ & $<0.0001$ & $<0.0001$ & $<0.0001$ & $<0.0001$ & $<0.0001$ \\

\modcfg{\pcirc{F}}{\pcirc{T1}}{\mcirc{T1c}}{\pcirc{T2}}
& 0.0203 & 0.0247 & 0.0031 & 0.0041 & 0.4209 & 0.4404 \\

\modcfg{\pcirc{F}}{\mcirc{T1}}{\pcirc{T1c}}{\pcirc{T2}}
& $<0.0001$ & $<0.0001$ & $<0.0001$ & $<0.0001$ & $<0.0001$ & $<0.0001$ \\

\modcfg{\mcirc{F}}{\pcirc{T1}}{\pcirc{T1c}}{\pcirc{T2}}
& $<0.0001$ & $<0.0001$ & $<0.0001$ & $<0.0001$ & $<0.0001$ & $<0.0001$ \\

\modcfg{\pcirc{F}}{\pcirc{T1}}{\pcirc{T1c}}{\pcirc{T2}}
& $<0.0001$ & $<0.0001$ & $<0.0001$ & $<0.0001$ & $<0.0001$ & $<0.0001$ \\
\hline
\end{tabular}}
}
\end{table*}
\begin{figure*}[t]
\includegraphics[width=\textwidth]{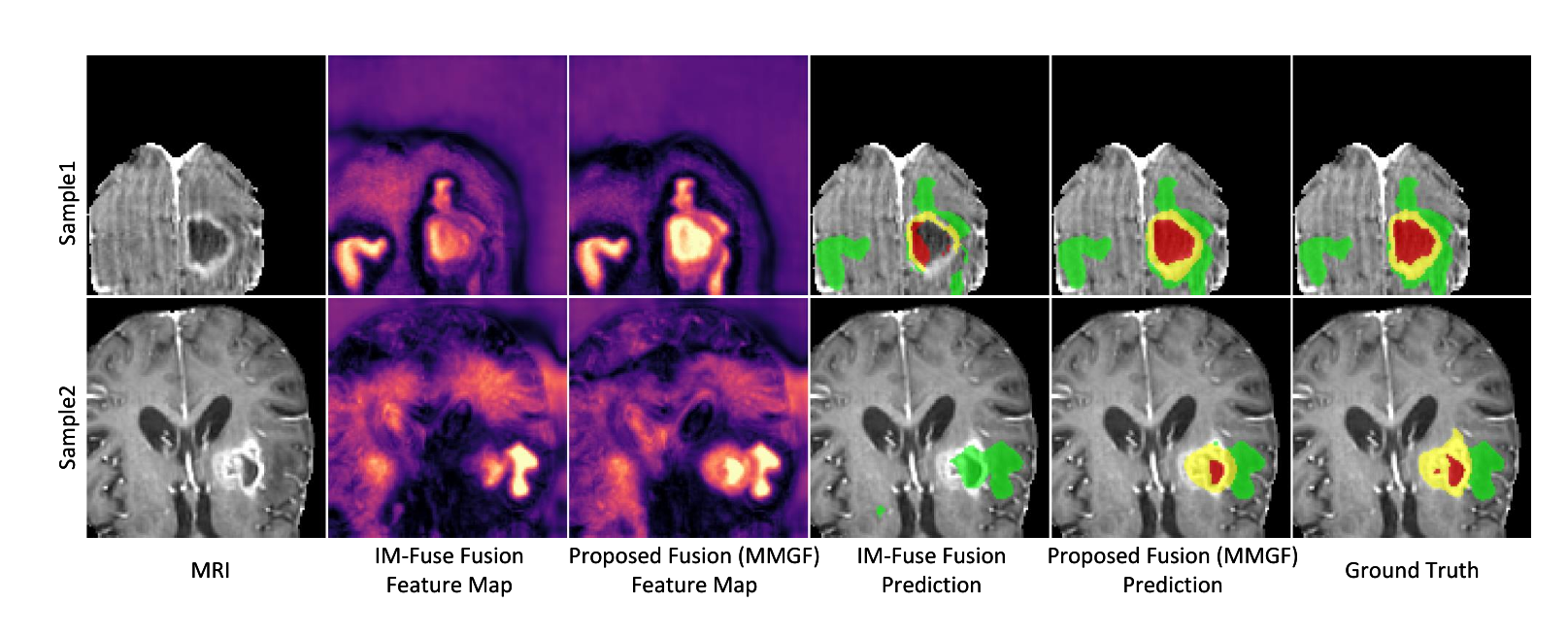}
\caption{Qualitative results for two representative test cases: BraTS-GLI-00322-000 (top) and BraTS-GLI-00737-000 (bottom), illustrating the impact of MMGF compared to IM-Fuse fusion on intermediate feature representations and final predictions. The green, yellow, and red colors in the predictions represent edema
(ED), enhancing tumor (ET), and necrotic core (NCR) respectively.}
 \label{fig:qual}
\end{figure*}
\subsection{Training Objective} 
The proposed segmentation network is trained 
using a combined Dice and cross-entropy loss to balance region overlap and voxel-wise classification. The error refinement module is supervised with a binary cross-entropy loss to identify erroneous or under-segmented regions, while the diffusion model is trained using a mean squared error objective in latent space.
\subsection{Results and Analysis}
Table~\ref{tab:brats2023_all} presents performance of the proposed method and its comparison with state of the art methods using BraTS 2023 Glioma dataset under different modality availability configurations for whole tumor (WT), tumor core (TC), and enhancing tumor (ET). For a fair comparison, all results are reported on the same BraTS 2023 test set which is used by IMFuse \cite{pipoli2025fuse}, which also benchmarks other recent approaches \cite{zhang2022mmformer,liu2023m3ae} under the same settings. 

The proposed method demonstrates consistently strong ET segmentation across all evaluated modality configurations. This behavior is clinically relevant, as ET is the most critical tumor sub-region and the most sensitive to missing contrast-enhanced information. For TC segmentation, the proposed method remains competitive across all modality configurations, while WT accuracy shows limited variation across methods, indicating that coarse tumor extent can be reliably recovered even under incomplete modality inputs.
We can also observe a consistent behavior across all evaluated methods from the results presented in Table~\ref{tab:brats2023_all}. In the presence of both FLAIR and T1ce modalities, segmentation accuracy remains comparable to the full-modality setting, even in the absence of T1, T2, or both. In contrast, removing T1ce leads to noticeable degradation in enhancing tumor (ET) and tumor core (TC) accuracy, whereas missing FLAIR primarily affects whole tumor (WT) segmentation, though to a lesser extent than the impact of missing T1ce.

Despite the observed sensitivity of all methods to missing T1ce and FLAIR information, the proposed method exhibits improved segmentation accuracy under missing-T1ce and FLAIR conditions. Compared to the recent state of the art IM-Fuse method \cite{pipoli2025fuse}, the proposed method achieves higher ET accuracy, with approximately 1.5-2.0\% absolute Dice improvements across multiple incomplete-modality configurations. For TC segmentation, the proposed method achieves 0.5-1.0\% Dice gains across several missing-modality settings and remains competitive in the remaining configurations. Importantly, these improvements are achieved with 2 million fewer parameters as compared to recent state of the art IM-Fuse method which has 47M. 

To further evaluate cross-cohort generalization, we tested the proposed method on a selected subset of 279 cases from the BraTS 2023 Meningioma cohort
under the same incomplete-modality settings, without any training or hyperparameter tuning etc. IM-Fuse and mmFormer methods were selected as comparative baselines because they demonstrated strong performance on the glioma benchmark dataset. All three methods were trained on the glioma cohort using the same experimental protocol, identical missing-modality configurations, and the same number of training epochs, and were subsequently evaluated on the meningioma test cases under the same evaluation protocol for fair comparison. 

The performance results for all the three methods on meningioma dataset are reported in Table~\ref{tab:brats2023_men}. The results show that the proposed method maintains robust performance under this cohort shift, with particularly consistent gains for the more challenging TC and ET regions as compared to the selected state of the art methods. These results indicate that the proposed method does not overfit to glioma-specific appearance, but instead learns generalized representations that remain effective when applied to a different tumor cohort. 
\begin{figure*}
\includegraphics[width=\textwidth]{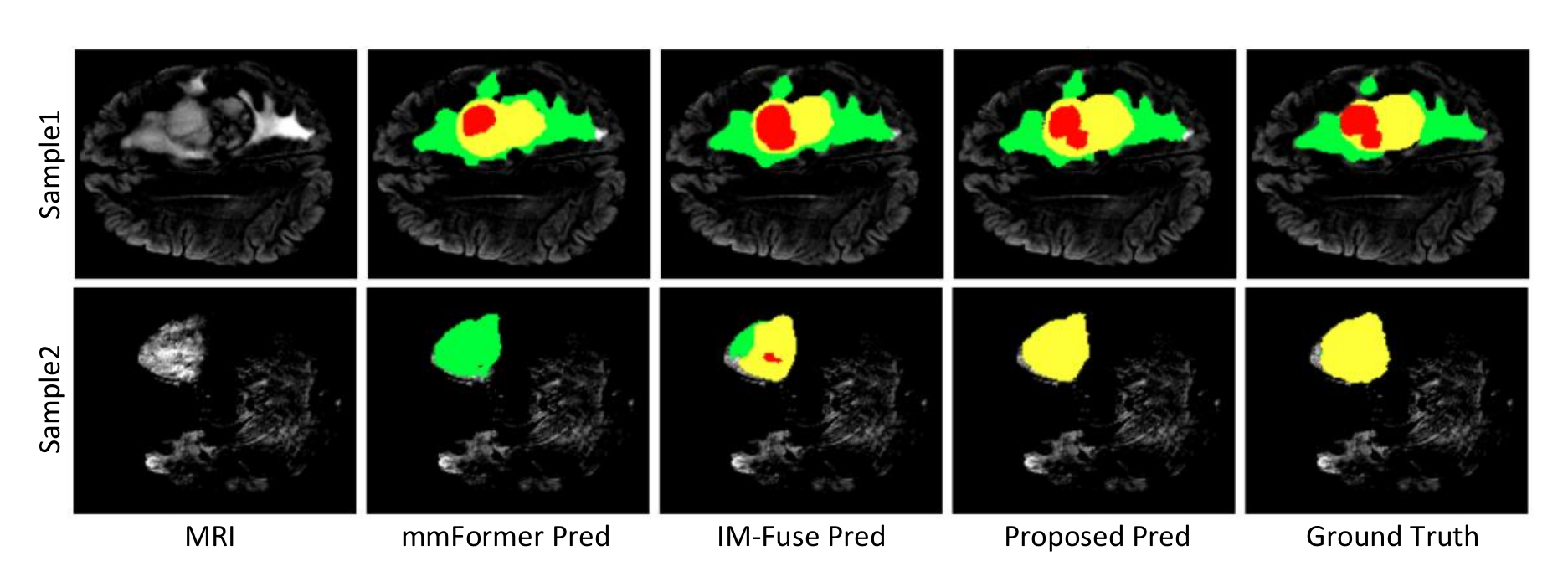}
\caption{Qualitative comparison of segmentation predictions for two representative BraTS 2023 Meningioma cases: BraTS-MEN-00669-000 (top) and BraTS-MEN-01009-000 (bottom). Columns show the input MRI, predictions from mmFormer, IM-Fuse, the proposed method, and the ground truth. The green, yellow, and red regions denote edema (ED), enhancing tumor (ET), and necrotic core (NCR), respectively.}
 \label{fig:qual_men}
\end{figure*}

To validate the statistical significance of cross-cohort improvements, we further performed paired two-sided Wilcoxon signed-rank tests against existing state of the art IM-Fuse method \cite{demvsar2006statistical}. Specifically, Dice scores from the proposed method and IM-Fuse were paired across the same test cases for each modality configuration and tumor region. Since the analysis involved 45 comparisons, corresponding to 15 modality configurations and three tumor regions (WT, TC, and ET), false discovery rate (FDR) correction was applied to the resulting $p$-values to account for multiple testing. 
The FDR-corrected p-value ($p_{\mathrm{FDR}}$) is reported in Table~\ref{tab:brats2023_full_stat_results}, showing statistically significant improvements across multiple missing-modality configurations and tumor regions, with several comparisons satisfying $p_{\mathrm{FDR}}<0.0001$. These results support the consistency of the observed improvements under cross-cohort incomplete-modality evaluation.

Beyond the quantitative improvements and subject-level statistical evidence, we further examine qualitative results to assess how the observed improvements translate into spatial segmentation quality.
As shown in 
Fig.~\ref{fig:qual},
the proposed MMGF improves both feature representations and final segmentation outputs. Compared with IM-Fuse-based fusion alone, which produces relatively diffuse activations within the tumor region, MMGF yields more localized and contrast-enhanced responses, particularly in the ET area. This difference is reflected in the final predictions, where MMGF achieves better ET delineation and closer alignment with the ground truth than IM-Fuse. The qualitative comparison 
on the meningioma test cases is presented in  Fig. \ref{fig:qual_men}, which further confirms the cross-cohort effectiveness of the proposed method. Compared with mmFormer and IM-Fuse, the proposed method provides closer agreement with the ground truth across all tumor sub-regions, where the baseline predictions show greater sub-region confusion.

\begin{table*}[t]
\centering
\caption{Ablation study under missing-T1ce (FLAIR, T1 available). 
}
\label{tab:ablation_t1ce}
\setlength{\tabcolsep}{12pt}
\resizebox{\textwidth}{!}{%
\begin{tabular}{l|cccc|ccc}
\hline
\multirow{2}{*}{Model} 
& \multicolumn{4}{c|}{Modules} 
& \multicolumn{3}{c}{Dice (\%)} \\
\cline{2-8}
& Mamba Fusion 
& MMGF 
& Diffusion 
& EGDR 
& WT 
& TC 
& ET \\
\cline{1-8}
\multirow{4}{*}{Proposed D$^3$Seg}  
& $\checkmark$ & $\times$ & $\times$ & $\times$ 
& 91.6 & 79.4 & 63.1 \\
 
& $\checkmark$ & $\checkmark$ & $\times$ & $\times$ 
& 92.2 & 79.7 & 64.4 \\
 
& $\checkmark$ & $\checkmark$ & $\checkmark$ & $\times$ 
& 92.6 & 80.3 & 65.5 \\
 
& $\checkmark$ & $\checkmark$ & $\checkmark$ & $\checkmark$ 
& \textbf{92.6} & \textbf{80.8} & \textbf{65.9} \\
\hline
\end{tabular}}
\end{table*}

\subsection{Ablation Study}
We conducted an ablation study on BraTS 2023 Glioma to evaluate the contribution of each component in the proposed $D^3$-Seg under a missing-T1ce setting (FLAIR and T1 available), as summarized in Table~\ref{tab:ablation_t1ce}. The model with only Mamba-based modality fusion achieves 91.6\% WT, 79.4\% TC, and 63.1\% ET Dice, with comparatively lower ET performance. Multi-hop Modality Graph Fusion results in 0.6\%, 0.3\% and 1.3\% improvements in WT, TC and ET Dice, respectively, over the Mamba-based fusion-only configuration. Diffusion-based T1ce feature imputation leads to an additional 0.4\%, 0.6\% and 1.1\% gain in WT, TC and ET Dice. The complete $D^3$-Seg model, including error-guided refinement, achieves the best performance (92.6\% WT, 80.8\% TC, 65.9\% ET), demonstrating the importance of these modules for improved segmentation under missing-modality conditions.
\begin{table*}[t]
\centering
\caption{Ablation of diffusion-based feature imputation under the available-T2 setting. T1ce and FLAIR denote diffusion-imputed latent features. Reliability-guided revision denotes remasking and correction of unreliable tokens during the diffusion sampling process. Inference time is reported per test sample.}
\label{tab:ablation_diffusion_imputation}
\setlength{\tabcolsep}{3pt}
\resizebox{\textwidth}{!}{%
\begin{tabular}{l|cc|c|ccc|cc}
\hline
\multirow{2}{*}{Variant}
& \multicolumn{2}{c|}{Imputed Features}
& \multirow{2}{*}{Reliability-guided Revision}
& \multicolumn{3}{c|}{Dice (\%)}
& \multicolumn{2}{c}{Efficiency}\\
\cline{2-3} \cline{5-9}
& T1ce & FLAIR
& 
& WT & TC & ET
& Time (s) & GFLOPs \\
\hline

No diffusion
& $\times$ & $\times$ & --
& 88.5 & 76.5 & 60.1
& 2.1 & 160 \\

T1ce imputation only
& $\checkmark$ & $\times$ & $\checkmark$
& 88.6 & 77.4 & 60.9
& 3.8 & 310 \\

FLAIR imputation only
& $\times$ & $\checkmark$ & $\checkmark$
& 88.7 & 77.0 & 60.4
& 3.8 & 310 \\

T1ce + FLAIR imputation
& $\checkmark$ & $\checkmark$ & $\checkmark$
& 88.7 & 77.7 & 61.2
& 4.7 & 420 \\

T1ce + FLAIR imputation
& $\checkmark$ & $\checkmark$ & $\times$
& 88.4 & 77.2 & 60.8
& 4.2 & 360 \\
\hline
\end{tabular}}
\end{table*}

\begin{table*}
\centering
\caption{Effect of reliability-guided revision under single-modality settings on BraTS 2023 Meningioma and BraTS 2023 Glioma. Results compare the proposed diffusion imputation module with and without remasking and revision during sampling. Dice Score (\%) is reported for WT, TC, and ET.}
\label{tab:ablation_revision_single_modality}
\setlength{\tabcolsep}{14pt}
\resizebox{\textwidth}{!}{%
\begin{tabular}{l|l|c|ccc}
\hline
\multirow{2}{*}{Dataset}
& \multirow{2}{*}{Available Modality}
& \multirow{2}{*}{Reliability-guided Revision}
& \multicolumn{3}{c}{Dice (\%)} \\
\cline{4-6}
& & & WT & TC & ET \\
\hline

\multirow{4}{*}{Glioma}
& T1 only & $\times$      & 82.2 & 76.1 & 58.7 \\
& T1 only & $\checkmark$ & 82.5 & 76.6 & 59.3 \\
& T2 only & $\times$      & 88.4 & 77.2 & 60.8 \\
& T2 only & $\checkmark$ & 88.7 & 77.7 & 61.2 \\
\hline

\multirow{4}{*}{Meningioma}
& T1 only & $\times$      & 74.5 & 64.4 & 47.5 \\
& T1 only & $\checkmark$ & 75.0 & 65.2 & 48.1 \\
& T2 only & $\times$      & 78.4 & 63.7 & 55.1 \\
& T2 only & $\checkmark$ & 78.7 & 64.1 & 55.6 \\
\hline

\end{tabular}}
\end{table*}

To further evaluate the impact of diffusion-based feature imputation, we conducted an additional ablation under the available-T2 setting, where the missing T1ce and FLAIR latent features are progressively imputed. As shown in Table~\ref{tab:ablation_diffusion_imputation}, the model without diffusion-based imputation achieves 88.5\% WT, 76.5\% TC, and 60.1\% ET Dice. Introducing T1ce feature imputation improves TC and ET Dice to 77.4\% and 60.9\%, respectively, indicating that recovering contrast-enhanced latent cues is particularly useful for tumor core and enhancing tumor delineation. In contrast, FLAIR feature imputation provides a slightly larger gain for WT, improving it from 88.5\% to 88.7\%, while also improving TC and ET over the no-diffusion baseline.

Joint T1ce+FLAIR feature imputation achieves the best overall performance, reaching 88.7\% WT, 77.7\% TC, and 61.2\% ET Dice. 
These results suggest that diffusion-based imputation contributes complementary missing-modality information. T1ce imputation mainly strengthens contrast-sensitive tumor sub-regions, whereas FLAIR imputation helps preserve broader tumor extent. 

The comparison between joint imputation with and without reliability-guided revision further shows that revising unreliable diffusion tokens improves segmentation accuracy from 88.4\%, 77.2\%, and 60.8\% to 88.7,\%, 77.7\% and 61.2\% for WT, TC and ET respectively at the cost of increased runtime and computational complexity. This indicates that reliability-guided revision improves the quality of imputed latent features, but introduces an accuracy--efficiency trade-off during inference.

To further validate the effectiveness of reliability-guided revision, we conduct an ablation study under single-modality input settings, where only T1 or only T2 is available. Results for both BraTS 2023 Glioma and BraTS 2023 Meningioma are presented in Table~\ref{tab:ablation_revision_single_modality}. Under these settings, the diffusion model imputes the missing T1ce and FLAIR latent representations from a single observed MRI sequence. Therefore, the imputed tokens need to preserve tumor-relevant information for downstream segmentation. However, when only a single modality is available, some imputed tokens may become less stable across diffusion sampling steps. Reliability-guided revision addresses this issue by remasking and correcting unreliable tokens during sampling. This targeted correction is reflected in the ablation results, where reliability-guided revision improves downstream segmentation accuracy across both datasets. On Glioma, it improves WT, TC, and ET Dice by 0.3\%, 0.5\%, and 0.4-0.6\% respectively. On Meningioma, the corresponding gains are 0.3-0.5\%, 0.4-0.8\%, and 0.5-0.6\% for WT, TC, and ET respectively.

\section{Conclusion}
We proposed a novel $D^3$-Seg model for brain tumor segmentation under missing-modality conditions. The proposed model comprises a new multi-hop modality graph fusion mechanism and a lightweight diffusion-based module to impute critical T1ce and FLAIR features, together with error-guided refinement to enhance tumor subregion delineation. Experiments on BraTS 2023 Glioma demonstrate consistent performance gains across WT, TC, and ET under diverse missing-modality configurations. To further assess generalization beyond the primary training cohort, we evaluated the trained model on a subset of the external BraTS 2023 Meningioma cohort, which was used exclusively for testing only. Results on this external cohort show that the proposed D$^3$Seg maintains superior segmentation performance compared with state-of-the-art methods under incomplete-modality settings. These findings highlight the effectiveness of combining graph-based fusion and modality feature imputation for improved segmentation with missing modalities.

\bibliographystyle{IEEEtran}
\bibliography{main}

\end{document}